\documentclass{article}

\usepackage{PRIMEarxiv}

\usepackage[utf8]{inputenc} 
\usepackage[T1]{fontenc}    
\usepackage{hyperref}       
\usepackage{url}            
\usepackage{booktabs}       
\usepackage{amsfonts}       
\usepackage{nicefrac}       
\usepackage{microtype}      
\usepackage{lipsum}
\usepackage{fancyhdr}       
\usepackage{graphicx}       
\graphicspath{{media/}}     
\usepackage{multirow}
\usepackage{amsmath}

\title{White-Box Diffusion Transformer for single-cell RNA-seq generation}

\author{
  Zhuorui Cui, Shengze Dong, Ding Liu\thanks{Corresponding author. E-mail: liuding@tiangong.edu.cn }\\
  School of Computer Science and Technology \\
  Tiangong University \\
  Tianjin, 300387\\
  People's Republic of China\\
}

\begin{document}
\maketitle

\begin{abstract}
As a powerful tool for characterizing cellular subpopulations and cellular heterogeneity, single cell RNA sequencing (scRNA-seq) technology offers advantages of high throughput and multidimensional analysis. However, the process of data acquisition is often constrained by high cost and limited sample availability. To overcome these limitations, we propose a model based on Diffusion model and White-Box transformer that aims to generate synthetic and biologically plausible scRNA-seq data. Diffusion model progressively introduce noise into the data and then recover the original data through a denoising process, a forward and reverse process that is particularly suitable for generating complex data distributions. White-Box transformer is a deep learning architecture that emphasizes mathematical interpretability. By minimizing the encoding rate of the data and maximizing the sparsity of the representation, it not only reduces the computational burden, but also provides clear insight into underlying structure. White-Box Diffusion Transformer combines the generative capabilities of Diffusion model with the mathematical interpretability of White-Box transformer. Through experiments using six different single-cell RNA-Seq datasets, we visualize both generated and real data using t-SNE dimensionality reduction technique, as well as quantify similarity between generated and real data using various metrics to demonstrate comparable performance of White-Box Diffusion Transformer and Diffusion Transformer in generating scRNA-seq data alongside significant improvements in training efficiency and resource utilization. Our code is available at \url{https://github.com/lingximamo/White-Box-Diffusion-Transformer}.

\end{abstract}


\section{Introduction}
Single-cell RNA sequencing (scRNA-seq), as an emerging sequencing technology, has become a powerful tool for describing cell subpopulation classification and cell heterogeneity by enabling high-throughput and multidimensional analysis of individual cells, thus avoiding the shortcomings of traditional sequencing in detecting the average transcriptional level of cell populations\cite{wang2009rna,wang2023evolution,marguerat2010rna}. However, obtaining scRNA-seq data is often constrained by high costs, limited sample availability, and the requirement for extensive laboratory resources. These limitations stimulate the application of generative models which are capable of producing synthetic, biologically plausible scRNA-seq data\cite{marouf2020realistic}. Generative Adversarial Networks (GANs)\cite{goodfellow2020generative} have been widely used to generate scRNA-Seq data, such as scIGANs\cite{xu2020scigans}, scGGAN\cite{huang2023scggan}, GRouNdGAN\cite{zinati2024groundgan}. However, the instability of training process and mode collapse of GANs may lead to certain issues, impacting the overall quality of the generated data. Variational Auto-Encoder(VAE)\cite{kingma2022autoencodingvariationalbayes} has also been used to analyze scRNA-seq data\cite{svensson2020interpretable}, such as scVAE\cite{gronbech2020scvae}, VASC\cite{wang2018vasc}, VEGA\cite{seninge2021vega}.

Diffusion model\cite{ho2020denoising} have recently garnered significant attention in generative modeling. They operate on the principle of gradually corrupting data with noise in a controlled manner, followed by a denoising process that recovers the original data. This forward and reverse process is particularly powerful for generating complex data distributions as it transforms the intricate structure of the data into a simpler Gaussian distribution which is then gradually refined back to its original complexity. Previous work proved that DDPM outperforms GAN in generation tasks\cite{dhariwal2021diffusion}. Although the Diffusion model is commonly applied in computer vision and natural language processing\cite{croitoru2023diffusion,chen2023diffusiondet,yang2023diffusion}, it has also found successful applications in bioinformatics\cite{guo2023diffusion}. For example, RoseTTAFold diffusion (RFdiffusion) is a framework for protein design that enables the design of diverse functional proteins from simple molecular specifications\cite{watson2023novo}. For generating scRNA-seq tasks, diffusion model also performs well, such as scRDiT\cite{dong2024scrdit}. Previous diffusion models have typically employed U-Net\cite{ronneberger2015u} as the noise predictor. Nevertheless, following the introduction of Diffusion Transformer (DiT), there has been a pronounced inclination towards adopting Transformer architectures for noise prediction tasks\cite{peebles2023scalable,gao2023masked}. This shift is motivated by the DiT's exceptional performance and improved scalability, which provide significant benefits in comparison to U-Net.

While Diffusion model excel in generating high-fidelity data, they often lack transparency and interpretability critical for model security issue. To address this issue, we turn to White-Box Transformer, which is a deep learning architecture emphasizing mathematical interpretability\cite{yu2024white,yu2023white,pai2023masked}. By leveraging sparse rate reduction principles, it minimizes the coding rate of data and maximizes the sparsity of the representation which not only reduces computational burden but also provides clear insight into underlying structures\cite{yu2020learningdiversediscriminativerepresentations,chan2020deep,chan2022redunet}. White-Box Transformer demonstrates practical effectiveness and is capable of learning to compress and sparsify representations of many large-scale real-world image and text datasets. Furthermore, it achieves performance levels that are remarkably close to those of transformer-based models such as ViT, MAE, DINO, BERT, and GPT2\cite{yu2024white,dosovitskiy2020image,he2022masked,zhang2022dino,devlin2018bert,radford2019language}. The advantages of White-Box Transformer include not only excellent performance, but also complete mathematical interpretability, relatively simple and fast models, and strong transfer learning capability.

In this paper, We introduce the White-Box Diffusion Transformer, a  model that integrates the strengths of both Diffusion model and White-Box Transformer. We briefly present the principles of Diffusion model and White-Box Transformer, detailing the architecture and the integration of the White-Box Transformer's components, such as Multi-Head Subspace Self-Attention (MSSA) and Iterative Shrinkage Thresholding Algorithm (ISTA), with the Diffusion process. Our experiments utilize six distinct single-cell RNA-Seq datasets, representing a diverse range of cell types and conditions. We employ t-SNE dimensionality reduction technique\cite{van2008visualizing} to visualize the generated and real data, providing a qualitative assessment of the model's performance. Additionally, we quantify the similarity between generated and real data using various metrics that showcase the model’s comparable performance to DiT in generating scRNA-seq data and significant improvements in training efficiency and resource utilization.

\section{Methods}
\label{sec:Methods}
In this section, we briefly present the principles of Diffusion model and White-Box Transformer model, and integrate them to create the White-Box Diffusion Transformer model, which combines their respective strengths to deliver superior performance and robust mathematical interpretability.

\subsection{Diffusion Model}
Diffusion model is a generative model that iteratively denoises and obtains target data samples from Gaussian noise. This process consists of two main steps: forward diffusion and backward inverse diffusion. During the forward diffusion, the model gradually introduces noise into the data until it is completely transformed. This can be viewed as a parameterized Markov chain. In the reverse process, the model predicts and removes noise to recover clear data samples.

The forward diffusion process of a diffusion model can be implemented through a series of iterative steps that gradually transform the data from its original distribution to a Gaussian distribution. Given a real sample $x_{0}\sim{q(x_{0})}$, where $q(\cdot)$ represents the distribution. Since each step $t$ of the forward process is only related to the forward step $t-1$, it can be regarded as a Markov process:

\begin{equation}
\begin{split}
    q(x_{T}|x_{0}) =\prod_{t=1}^Tq(x_{t}|x_{t-1}), \quad
    q(x_{t}|x_{t-1}) =\mathcal{N}(x_{t};\sqrt{1-\beta_{t}}x_{t-1}, \beta_{t}I)
\end{split}
\end{equation}

Reparameterization is used to represent the data distribution at each step of the forward process:

\begin{equation}
    x_{t}=\sqrt{\alpha_{t}}x_{t-1}+\sqrt{1-\alpha_{t}}\epsilon
\end{equation}

where $x_{t}$ represents the data at diffusion step $t$, $x_{t-1}$ is the data from the previous step, $\epsilon$ is a Gaussian noise sampled from a standard normal distribution $\mathcal{N}(0,I)$, and $\alpha_{t}$ is a coefficient less than one that decreases over time $t$, controlling the amount of noise added at each step, $\alpha_{t}$ is defined as:

\begin{equation}
    \alpha_{t}=1-\beta_{t}
\end{equation}

where $\beta_{t}$ is a directly set hyperparameter that typically increases with the time step t and is utilized to regulate the variance of the noise at each step. In practice, $\beta_{t}$ gradually increases from $\beta_{1}$ to $\beta_{T}$, which ensures that the data gradually tends to a pure Gaussian noise distribution as the diffusion step proceeds.

Above process can be viewed as gradually introducing noise into the data until it is completely transformed into noise. Furthermore, to directly calculate $x_{t}$ from the initial data $x_{0}$ without the need for step-by-step iteration, the following formula can be used:

\begin{equation}
\label{eq:one_step_add_noise}
    x_{t}=\sqrt{\bar{\alpha_{t}}}x_{0}+\sqrt{1-\bar{\alpha_{t}}}\epsilon, \ \ \mathrm{where} \ \ \bar{\alpha_{t}}=\prod_{i=1}^t\alpha_{i}
\end{equation}

The key to the forward diffusion process is the gradual increase of noise until the data is entirely converted into Gaussian noise. This process can be viewed as a Markov chain, where each step depends only on the state of the previous step. In this way, the diffusion model can gradually simplify complex data distributions to a simple Gaussian distribution, providing a foundation for the subsequent reverse diffusion process, that is, data generation or recovery.

The reverse diffusion process of a diffusion model, also known as the denoising process, is the procedure of recovering the original data from the noisy data obtained through the forward diffusion process. This can be achieved by a series of iterative steps, where each step is designed to remove a certain amount of noise, culminating in the acquisition of clear data samples.

If the reversed distribution $q(x_{t-1}|x_{t})$ can be progressively obtained, it is possible to reconstruct the original distribution $x_{0}$ from the completely standard Gaussian noise distribution $x_{T}\sim{\mathcal{N}(0,I)}$. If $q(x_{t}|x_{t-1})$ satisfies a Gaussian distribution and $\beta_{t}$ is sufficiently small, $q(x_{t-1}|x_{t})$ remains a Gaussian distribution. However, it is impossible to simply determine $q(x_{t-1}|x_{t})$, hence we employ a deep learning model with parameters $\theta$ to predict such a reverse distribution $p_{\theta}$:

\begin{equation}
    p_{\theta}(x_{0}|x_{T})=\prod_{t=1}^Tp_{\theta}(x_{t-1}|x_{t})
\end{equation}

\begin{equation}
\label{eq:6}
    p_{\theta}(x_{t-1}|x_{t})=\mathcal{N}(x_{t-1};\mu_{\theta}(x_{t},t),\Sigma_{\theta}(x_{t},t)),\ \  \mathrm{where} \ \  \Sigma_{\theta}(x_{t},t)=\sigma_{t}^2I.
\end{equation}

Although we cannot obtain the reversed distribution $q(x_{t-1}|x_{t})$ directly, if we know $x_{0}$, we can derive $q(x_{t-1}|x_{t},x_{0})$ using Bayesian formula as follows:

\begin{equation}
    q(x_{t-1}|x_{t},x_{0})=\mathcal{N}(x_{t-1};\tilde{\mu}(x_{t},x_{0}),\tilde{\beta_{t}}I)
\end{equation}

where

\begin{equation}
\label{eq:7}
    \tilde{\mu}(x_{t},x_{0})=\frac{\sqrt{\bar{\alpha}_{t-1}}\beta_{t}}{1-\bar{\alpha_{t}}}x_{0}+\frac{\sqrt{\alpha_{t}}(1-\bar{\alpha}_{t-1})}{1-\bar{\alpha_{t}}}x_{t}
\end{equation}

\begin{equation}
    \tilde{\beta_{t}}=\frac{1-\bar{\alpha}_{t-1}}{1-\bar{\alpha_{t}}}\beta_{t}
\end{equation}

From Eq. \eqref{eq:one_step_add_noise}, it can be deduced that $x_{0}=\frac{1}{\sqrt{\bar{\alpha_{t}}}}(x_{t}-\sqrt{1-\bar{\alpha_{t}}}\epsilon)$. Substituting this into Eq. \eqref{eq:7} yields:

\begin{equation}
\label{eq:10}
    \mu_{\theta}(x_{t},t)=\frac{1}{\sqrt{\alpha_{t}}}(x_{t}-\frac{1-\alpha_{t}}{\sqrt{1-\bar{\alpha}_t}}\epsilon_{\theta}(x_t,t))
\end{equation}

where $\epsilon_{\theta}(x_t,t)$ represents the Gaussian noise predicted by the deep neural network model from $x_t$. It can be deduced from Eq. \eqref{eq:6} and \eqref{eq:10} that to sample $x_{t-1}\sim{p_{\theta}(x_{t-1}|x_{t})}$ is to compute:

\begin{equation}
    x_{t-1}=\frac{1}{\sqrt{\alpha_{t}}}(x_{t}-\frac{1-\alpha_{t}}{\sqrt{1-\bar{\alpha}_t}}\epsilon_{\theta}(x_t,t))+\sigma_{t} z,\ \ \mathrm{where} \ \ z\sim{\mathcal{N}(0,I)}
\end{equation}

Experimentally, both $\sigma_{t}^2=\beta_{t}$ and $\sigma_{t}^2=\bar{\beta_{t}}$ had similar results\cite{ho2020denoising}, so we choose $\sigma_{t}^2=\beta_{t}$.

\subsection{White-Box Transformer}
\label{White-Box Transformer}
White-Box Transformer is an innovative deep learning network architecture designed to learn compressed and sparse representations of data through the principle of sparse rate reduction\cite{yu2024white,yu2020learningdiversediscriminativerepresentations}. This architecture is distinguished by its complete mathematical interpretability, that is, the operations of each network layer can be described with mathematical formulas that are consistent with their design objectives at a hierarchical level.

Sparse rate reduction is a principle to measuring the quality of representations, aiming to minimize the coding rate of data while maximizing the sparsity of the representation. The optimization goal of this principle can be articulated as follows:

\begin{equation}
    \mathop{\max}_{f\in F}E_{Z=f(X)}\big[R(Z)-R^c(Z \mid U_{[K]})-\lambda \Vert Z \Vert_0\big]
\end{equation}

where $R(Z)$ is the lossy coding rate of the feature set $Z\in R^{d\times n}$, representing the total amount of information of the feature set. $R^c(Z \mid U_{[K]})$ is the conditional lossy coding rate of the feature set under the given subspaces set $U_{[K]}$, reflecting the consistency between the feature set and the subspace structure. $\Vert Z \Vert_0$ is the $l_{0}$ norm of $Z$, that is, the number of non-zero elements in $Z$, used to promote sparsity. $\lambda$ is the regularization parameter used to balance the encoding rate and sparsity. And orthonormal bases $U_{[K]}=(U_{k})_{k=1}^K\in (R^{d\times p})^K$, which comprises $K$ randomly generated Gaussian distributions. $n$ is the number of samples and $\epsilon$ is the given rate distortion. $R(Z)$ and $R^c(Z \mid U_{[K]})$ are defined as follows\cite{ma2007segmentation}:

\begin{equation}
\label{eq:13}
    R(Z)=\frac{1}{2}\mathrm{logdet}(I+\alpha Z^*Z), \quad \mathrm{where} \ \  \alpha=\frac{d}{n\epsilon^2}
\end{equation}

\begin{equation}
    R^c(Z \mid U_{[K]})=\sum_{k=1}^KR(U_{k}^*Z)=\frac{1}{2}\sum_{k=1}^K\mathrm{logdet}(I+\beta(U_{k}^*Z)^*(U_{k}^*Z)),\quad \mathrm{where} \ \ \beta=\frac{p}{n\epsilon^2}
\end{equation}

The main architecture of the White-Box Transformer consists of multiple layers stacked on top of each other, with each layer including a compression step and a sparsification step:

\begin{equation}
    Z^l \xrightarrow{\mathrm{MSSA}} Z^{l+1/2} \xrightarrow{\mathrm{ISTA}} Z^{l+1}
\end{equation}

where $Z^{l+1/2}$ is chosen to incrementally minimize $R^c(Z^{l+1/2}\mid U_{[K]}^l)$ and $Z^{l+1}$ is chosen to incrementally minimize $\big[\lambda\Vert Z^{l+1} \Vert_0 -R(Z^{l+1})\big]$\cite{yu2024white}.

The design goal of Multi-Head Subspace Self-Attention (MSSA) is to reduce the coding rate of feature representations through compression operations while preserving the important information in the data. This operation is achieved by implementing a self-attention mechanism across multiple subspaces, with each subspace corresponding to an attention head. MSSA reduces the coding rate $R^c$ through an approximate gradient descent step. Given the input $Z^{l}$, the output $Z^{l+1/2}$ of MSSA is given by the following formula:

\begin{equation}
    Z^{l+1/2}=Z^l+\mathrm{MSSA}(Z^l\mid U_{[K]}^l)\approx Z^l-\eta \nabla R^c(Z^l\mid U_{[K]}^l)
\end{equation}

where $\eta$ is the learning rate, $\nabla R^c(Z^l\mid U_{[K]}^l)$ is the gradient of the coding rate with respect to the features $Z$, and $U_{[K]}^l=\sum_{k=1}^KU_{k}^l$.

For each subspace $U_{k}$, the Subspace Self-Attention (SSA) computes similarity through the projection of the input features $Z$ onto that subspace and converts it to a distribution of membership through a softmax function\cite{yu2024white}. SSA operator resembles the attention operator in a typical transformer\cite{vaswani2023attentionneed}, SSA is defined as:

\begin{equation}
    \mathrm{SSA}(Z\mid U_{k})=(U_{k}^*Z)\mathrm{softmax}((U_{k}^*Z)^*(U_{k}^*Z)),\ \ k\in [K]
\end{equation}

MSSA aggregates information by combining the outputs of SSA from all subspaces. If there are $K$ subspaces, the output of MSSA can be represented as:

\begin{equation}
    \mathrm{MSSA}(Z\mid U_{[K]})=\beta \big[U_{1},...,U_{k}\big]\begin{bmatrix}\mathrm{SSA}(Z\mid U_{1})\\ \vdots \\ \mathrm{SSA}(Z\mid U_{k})\end{bmatrix}
\end{equation}

The design of MSSA allows the model to independently learn representations in different subspaces, which helps to capture the multidimensional structure of the data. By optimizing the coding rate, MSSA promotes the compactness of the feature representation while maintaining the intrinsic complexity of the data.

Iterative Shrinkage Thresholding Algorithm (ISTA) is used for sparsifying feature representations\cite{beck2009fast}. This is achieved by solving a regularized optimization problem aimed at minimizing the encoding rate while increasing the sparsity of the representation.

Assuming a complete and orthogonal global dictionary $D^l\in R^{d\times d}$, and ask to sparsify the intermediate iterates $Z^{l+1/2}$ with respect to $D^l$\cite{yu2024white}. In mathematical terms, the expression is $Z^{l+1/2}=D^lZ^{l+1}$, where $D^{l+1}$ is more sparse. From Eq. \eqref{eq:13} and previous statement that $Z^{l+1}$ is chosen to incrementally minimize $\big[\lambda\Vert Z^{l+1} \Vert_0 -R(Z^{l+1})\big]$, we can deduce:

\begin{equation}
    Z^{l+1}\approx \mathop{\mathrm{arg\ min}}_{Z}\Vert Z \Vert_0 \quad \mathrm{subject\ to} \quad Z^{l+1/2}=D^lZ
\end{equation}

This optimization problem can be easily transformed into an unconstrained convex optimization problem using LASSO\cite{wright2022high}, with the addition of a positive constraint:

\begin{equation}
    Z^{l+1}\approx \mathop{\mathrm{arg\ min}}_{Z\ge 0}\big[\lambda\Vert Z \Vert_1 +\frac{1}{2}\Vert Z^{l+1/2}-D^lZ \Vert_F^2 \big]
\end{equation}

Here, $\Vert Z \Vert_1$ represents the $l_{1}$ norm of $Z$, which promotes sparsity, and $\Vert \cdot \Vert_F$ denotes the Frobenius norm, used to measure the discrepancy between $Z^{l+1/2}$ and $D^lZ$.

ISTA updates $Z$ iteratively, with each iteration consisting of a gradient descent step and a nonlinear activation function, such as the Rectified Linear Unit (ReLU)\cite{agarap2019deeplearningusingrectified}, to enforce sparsity. The update rule is as follows:

\begin{equation}
    Z^{l+1}=\mathrm{ISTA}(Z^{l+1/2}\mid D^l),\quad \mathrm{ISTA}(Z\mid D)=\mathrm{ReLU}(Z-\eta D^\ast(DZ-Z)-\eta \lambda_{1})
\end{equation}

\subsection{White-Box Diffusion Transformer}
We introduce a novel deep learning framework that integrates the Diffusion Transformer (DiT)\cite{peebles2023scalable} with the White-Box Transformer, creating a hybrid model that capitalizes on the strengths of both. At its core, our framework utilizes the fully mathematically interpretable White-Box Transformer as the noise predictor for DiT. This synergy not only preserves the high quality of generated samples and excellent generalization capabilities but also achieves high mathematical interpretability and efficiency, thus materializing a White-Box Diffusion Transformer.

\begin{figure}[htbp]
    \label{fig:network}
    \centering
    \includegraphics[width=1\textwidth]{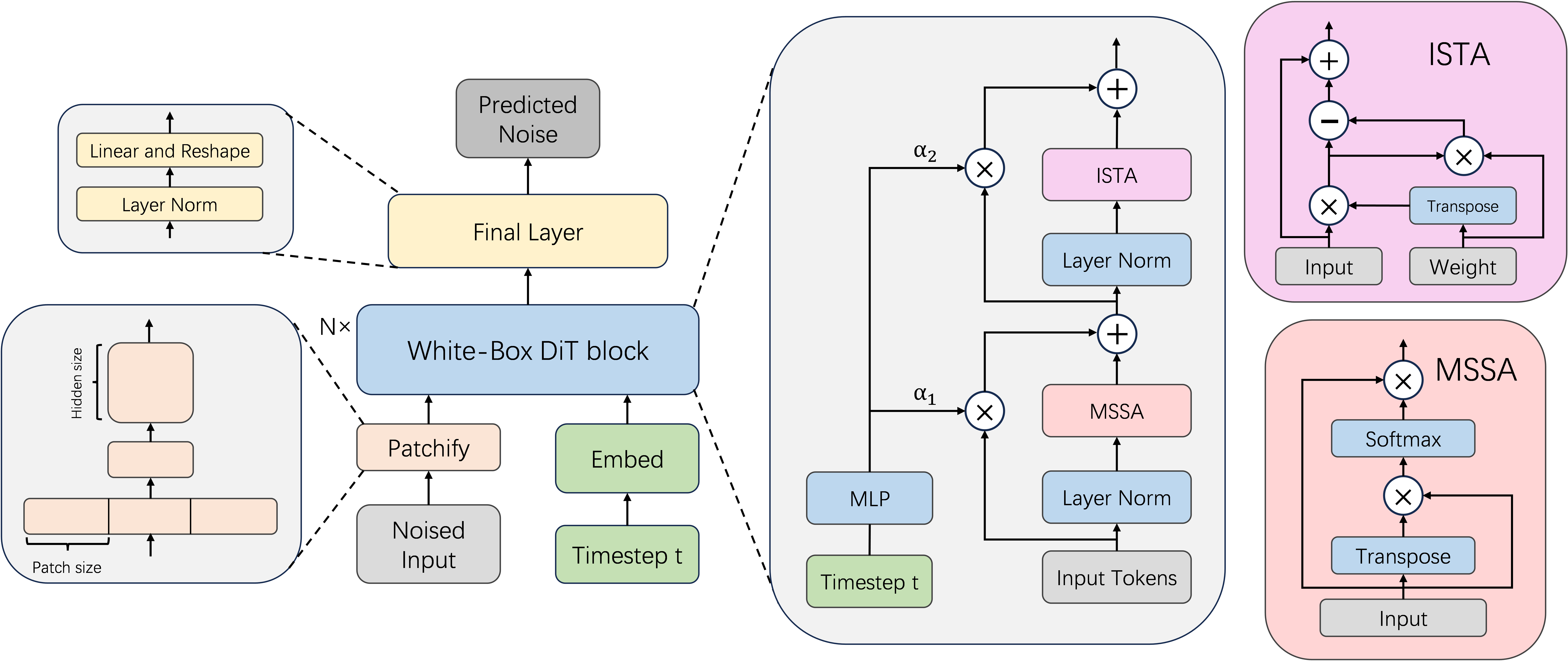}
    \caption{Architecture of noise predictor for White-Box Diffusion Transformer. We employ the MSSA layer for data compression, replacing the Attention layer. Additionally, we utilize the ISTA layer to achieve data sparsity and replace the FFN layer.}
\end{figure}

As outlined in section \ref{White-Box Transformer}, the essence of the White-Box Transformer lies in its data compression through the Multi-Head Subspace Self-Attention (MSSA) layer, followed by data sparsification via the Iterative Shrinkage Thresholding Algorithm (ISTA) layer. Aligning with the functionalities of the Multi-Head Self-Attention and Pointwise Feedforward mechanisms, we have integrated the MSSA and ISTA layers into DiT, as illustrated in Figure \ref{fig:network}.

We input scRNA-seq data with Gaussian noise and a time step $t$ into the model, which then passes through $N$ × White-Box DiT blocks and a final layer to obtain the predicted noise. For instance, in the case of a scRNA-seq sample with $n$ genes, the input tensor size is $(n,1)$. The Patchify layer segments the tensor into $n/p$ patches and converts these patches into a sequence of tokens with a hidden dimension $d$. Here, $p$ and $d$ are hyperparameters that can be adjusted to modify the size and a smaller value for $p$ leads to higher computational complexity due to increased token processing requirements. The Patchify layer outputs a tensor $X$ of size $(n/p,d)$ and after embedding the time step $t$ into a tensor $T$ of the same dimension, both $X$ and $T$ are fed into $N$ × White-Box DiT blocks, followed by the final layer, yielding the model's predicted noise output. Additionally, the "Weight" in the ISTA block refers to the matrix obtained after kaiming initialization.

\section{Results}
\label{sec:results}
In this section, we conduct experiments to study the performance of DiT and White-Box DiT by generating single cell RNA-seq samples of the six datasets. When we employed White-Box DiT for training, we observed its strong capability in fitting scRNA-seq data and demonstrating robustness. Upon comparison with DiT, we found that both White-Box DiT and DiT yield similar data generation effects, with slight improvements in certain indicators over DiT. In essence, White-Box DiT and DiT exhibit comparable performance. However, the White-Box DiT model significantly reduces time overhead during training and data generation processes while requiring substantially fewer resources.

\begin{figure}[htbp]
    \centering
    \includegraphics[width=1\textwidth]{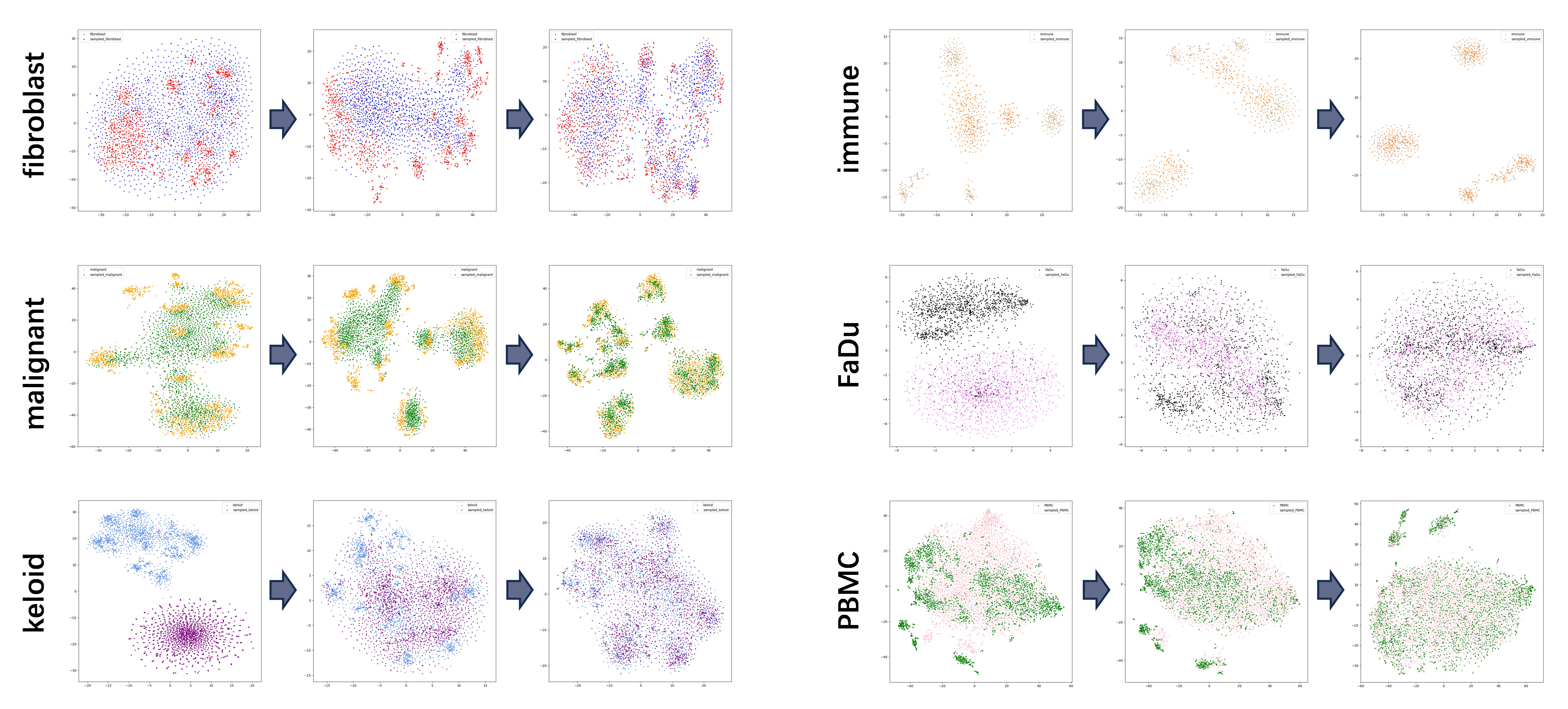}
    \caption{Fitting process of White-Box Diffusion Transformer. Scatter plots of the data generated by the White-Box Diffusion Transformer model with varying training epochs and real data are generated after dimensionality reduction using t-SNE.}
    \label{fig:training}
\end{figure}

\subsection{Results of single cell RNA-seq generation}
In order to substantiate the feasibility of the White-Box Diffusion Transformer, we conducted training experiments utilizing six distinct single-cell RNA-seq datasets. These six single-cell RNA-seq datasets were fibroblast, malignant, keloid, immune, FaDu and PBMC. The fibroblast dataset and the maglinant dataset were from a tumor cell dataset (GSE103322), which contained 5902 oral cavity head and neck squamous cell carcinomas cells from 18 patients\cite{puram2017single}. The keloid dataset (GSE181297) consisted of 1936 keloid cells\cite{Shim2022} while the immune dataset (GSE103322) encompassed 407 immune cells\cite{puram2017single}. Deriving from the Single-cell atlas of human cell lines dataset\cite{zhu2023single}, the FaDu dataset contained 1374 human pharyngeal squamous cell carcinoma cells. Considering the constraints of available computing power, we implemented a random sampling strategy, selecting 3,000 cells from the original Peripheral Blood Mononuclear Cells (PBMC) dataset\cite{kleiveland2015peripheral} to constitute a new dataset for training.

To achieve a visual representation of the generated outcomes, we utilized the t-SNE dimensionality reduction algorithm\cite{van2008visualizing} to consistently map the real and synthetic data onto a two-dimensional space, facilitating the generation of a scatter plot for analysis. Employing this visualization technique, we presented a scatter plot shown in Figure \ref{fig:training}, which visually represents the sample and real data generated by the White-Box Diffusion Transformer at different training epochs. The results show that as the number of training epochs increases, the distribution of the sample data generated by the White-Box Diffusion Transformer progressively aligns with that of the real data distribution.

To quantitatively assess the discrepancy between generated data and real data, we computed the Kullback-Leibler divergence (KL divergence)\cite{kullback1951information}, Wasserstein distance\cite{kullback1951information}, and Maximum Mean Discrepancy (MMD)\cite{gretton2006kernel}, as detailed in Table \ref{tab:epoch-comparison}. The table presents the quality metrics of the data generated by the White-Box Diffusion Transformer across various training epochs.

\begin{table}[htpb]
\renewcommand\arraystretch{1.2}
\centering
\scriptsize
\caption{Quality of data generated by the White-Box Diffusion Transformer model on six datasets. KL divergence, Wasserstein distance and MMD are used to measure the quality of the generated data for different training epochs, respectively.}
\label{tab:epoch-comparison}

\resizebox{0.8\textwidth}{!}{
\begin{tabular}{c|c|ccc}
\hline
Method                      & \makebox[0.07\textwidth]{epoch} & \makebox[0.15\textwidth]{KL Divergnece} & \makebox[0.15\textwidth]{Wasserstein Distance} & \makebox[0.15\textwidth]{MMD}    \\ \hline
\multirow{3}{*}{fibroblast} & 400   & 4.2366        & 0.0385               & 0.0310 \\
                            & 800  & 4.4631        & 0.0644               & 0.0389 \\
                            & 1200  & 4.3079        & 0.0178               & 0.0223 \\ \hline
\multirow{3}{*}{malignant}  & 800   & 3.2813        & 0.0422               & 0.0636 \\
                            & 1600   & 3.3351        & 0.0576               & 0.0698 \\
                            & 2400  & 3.2521        & 0.0227               & 0.0255 \\ \hline
\multirow{3}{*}{keloid}     & 600   & 0.1320        & 0.0099               & 0.0671 \\
                            & 800   & 0.1426        & 0.0124               & 0.0893 \\
                            & 1200  & 0.1491        & 0.0046               & 0.0140 \\ \hline
\multirow{3}{*}{immune}     & 800   & 7.2587        & 0.0481               & 0.0504 \\
                            & 1200  & 7.3595        & 0.0379               & 0.0355 \\
                            & 2400  & 7.3149        & 0.0158               & 0.0284 \\ \hline
\multirow{3}{*}{FaDu}       & 400   & 1.1635        & 0.0783               & 0.0493 \\
                            & 800   & 1.1693        & 0.0617               & 0.0375 \\
                            & 1200  & 1.2144        & 0.0377               & 0.0182 \\ \hline
\multirow{3}{*}{PBMC}       & 800   & 6.9464        & 0.1141               & 0.0617 \\
                            & 1600   & 6.7255        & 0.0831               & 0.0318 \\
                            & 2000  & 6.6757        & 0.0542               & 0.0154 \\ \hline
\end{tabular}
}
\end{table}

White-Box Diffusion transformer is robust. For each data category, we employed the best-performing model to generate a dataset that is five times larger than the real data samples. Subsequently, t-SNE dimensionality reduction technique was applied to create a visual scatter plot, as depicted in Figure \ref{fig:mix}-b. The results indicate that the generation of large-scale samples does not compromise the quality of the synthetic data, which shows a strong consistency with the real data, thereby substantiating the robustness and stability of our model.


\subsection{Comparsion of White-Box DiT and DiT}
The quality of the generated data is a critical measure for evaluating a model's performance, but it is not the only one. We evaluated the performance of both the White-Box DiT and DiT through training and data generation on six single-cell RNA-seq datasets.

\begin{table}[htbp]
\renewcommand\arraystretch{1.3}
\centering
\caption{Generation results of White-Box DiT model and DiT model on six single-cell RNA-seq datasets. We use KL divergence, Wasserstein distance and MMD to measure the quality of the data generated by the two models. The data presented in the table represent the optimal metrics attained upon completion of model training.}
\label{tab:network-comparison}

\resizebox{\textwidth}{!}{
\begin{tabular}{c|ccc|ccc}
\hline
Model      & \multicolumn{3}{c|}{White-Box DiT}            & \multicolumn{3}{c}{DiT}                       \\ \hline
Type     & \makebox[0.15\textwidth]{KL Divergnece} & \makebox[0.15\textwidth]{Wasserstein Distance} & \makebox[0.14\textwidth]{MMD}    & \makebox[0.15\textwidth]{KL Divergnece} & \makebox[0.15\textwidth]{Wasserstein Distance} & \makebox[0.14\textwidth]{MMD}    \\ \hline
fibroblast & 3.8735        & 0.0120               & 0.0223 & 3.9717        & 0.0269               & 0.0229 \\
malignant  & 3.2521        & 0.0068               & 0.0255 & 3.1379        & 0.0076               & 0.0469 \\
keloid     & 0.1374        & 0.0046               & 0.0140 & 0.1515        & 0.0054               & 0.0152 \\
immune     & 6.8828        & 0.0158               & 0.0284 & 7.0978        & 0.0128               & 0.0303 \\
FaDu       & 1.1635        & 0.0377               & 0.0182 & 1.0999        & 0.0527               & 0.0264 \\
PBMC       & 6.6353        & 0.0409               & 0.0154 & 6.6273        & 0.0312               & 0.0122 \\ \hline
\end{tabular}
}
\end{table}

\begin{figure}[htbp]
    \centering
    \includegraphics[width=1\textwidth]{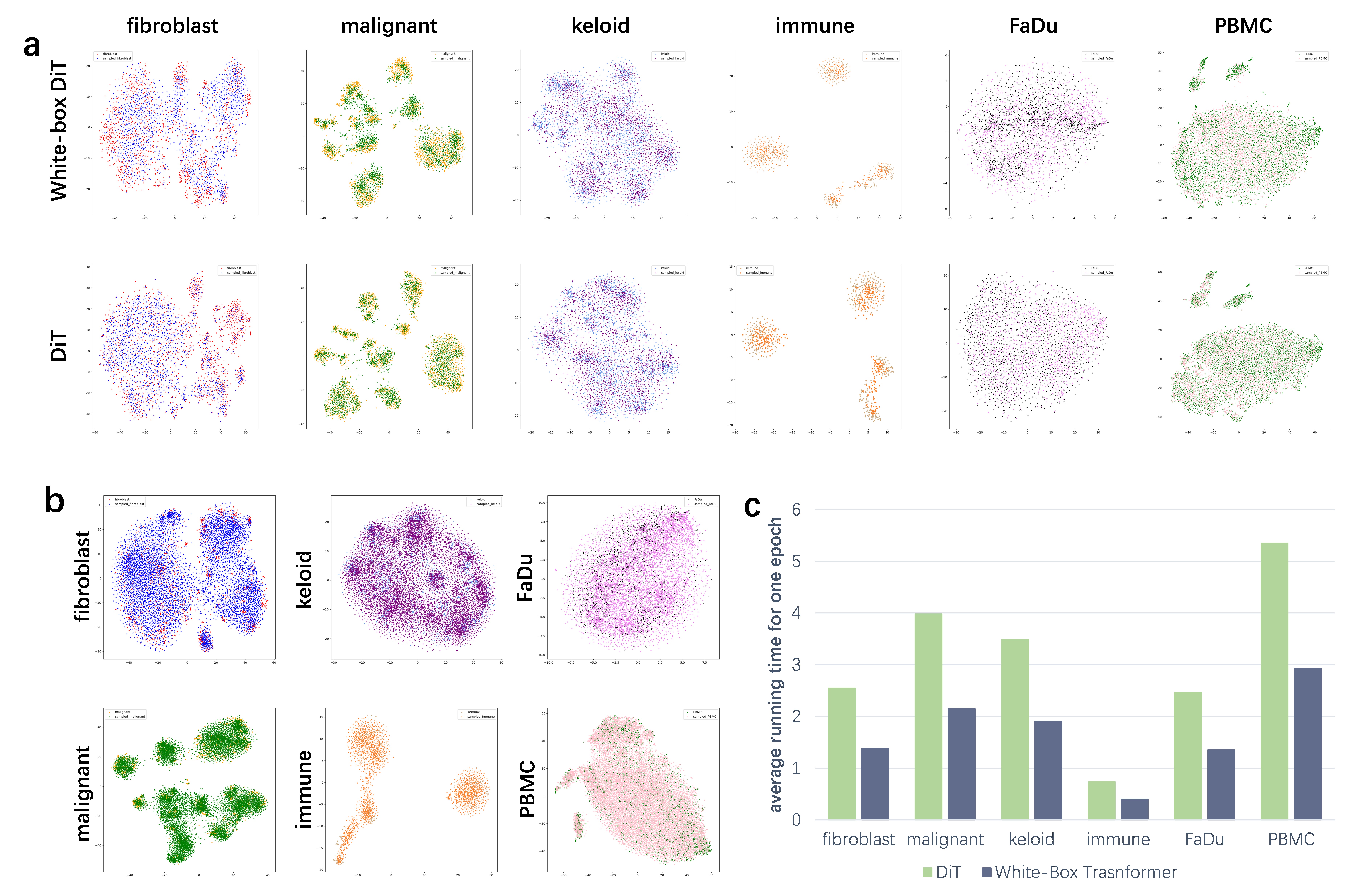}
    \caption{\textbf{a.}  Scatter plots of generated and real data after t-SNE dimensionality reduction for White-Box DiT and DiT.  \textbf{b.} Scatter plots of large batches of generated data and actual data following t-SNE dimensionality reduction.\textbf{c.} Average running time per epoch when training DiT and White-Box DiT.}
    \label{fig:mix}
\end{figure}

White-Box DiT achieves similar performance as DiT in data generation. Figure \ref{fig:mix}-a presents a scatter plot illustrating the single cell RNA-seq data generated by the two models following t-SNE dimensionality reduction. From this visualization, it is apparent that the synthetic data produced by both models closely resemble the actual data. Furthermore, as depicted in Table \ref{tab:network-comparison}, the KL divergence, Wasserstein distance, and MMD computed from the generated data indicate a high degree of similarity between the outputs of the two models. Notably, the data generation quality of the White-Box DiT is marginally superior to that of the DiT on most datasets.

White-Box DiT is faster to train and consumes less amount of resources than DiT. Upon training on an NVIDIA RTX A6000 GPU, we recorded the average running time for each model across six datasets, as depicted in Figure \ref{fig:mix}-c. The results indicate that White-Box DiT consistently achieves approximately half the average running time per training epoch compared to DiT. Within the same device memory constraints, White-Box DiT is capable of stacking a greater number of layers and more heads of Muti-Head Subspace Self-Attention(MSSA), which potentially leads to improved performance. When comparing the two models under the same depth and heads configurations, White-Box DiT is observed to save checkpoints in a much smaller size than DiT, where DiT is 129.81MB and white-box DiT is 68.98MB. It suggests that White-Box DiT could substantially decrease the hard disk storage requirements.


White-Box DiT generates samples more efficiently than DiT. DiT's data generation process is notably slow, prompting us to utilize the DDIM\cite{song2020denoising} sampling approach to accelerate sample generation and reduce time overhead. Also, it is observed that employing DDIM sampling approach with White-Box DiT further minimizes the time overhead. For instance, when using 10x acceleration sampling, DiT consumes 2.13 minutes when generating 2215 malignant data, while White-Box DiT only takes 1.18 minutes. 

\section{Conclusion}
In this paper, we propose the White-Box Diffusion Transformer, a framework that combines the mathematical interpretability of the White-Box Transformer with the generative capabilities of the Diffusion Transformer. Employing the White-Box Transformer as a noise predictor, we have trained a model for the generation of scRNA-seq data, outperforming several other neural networks. Our model demonstrates an exceptional capability to reconstruct scRNA-seq data from complete Gaussian noise, yielding results that is similar to the original data. The model provides a feasible solution to solve the problem of limited scRNA-seq samples. Our experimental results show that compared with DiT, White-Box Diffusion Transformer has distinct advantages in improving data generation efficiency and reducing time overhead, while generating samples with similar quality.


\section*{Acknowledgments}
DL is supported by the Tianjin Natural Science Foundation of China (20JCYBJC00500), the Science \& Technology Development Fund of Tianjin Education
Commission for Higher Education (2018KJ217).

\bibliographystyle{unsrt}  
\bibliography{references}

\begin{thebibliography}{10}

\bibitem{wang2009rna}
Zhong Wang, Mark Gerstein, and Michael Snyder.
\newblock Rna-seq: a revolutionary tool for transcriptomics.
\newblock {\em Nature reviews genetics}, 10(1):57--63, 2009.

\bibitem{wang2023evolution}
Shuo Wang, Si-Tong Sun, Xin-Yue Zhang, Hao-Ran Ding, Yu~Yuan, Jun-Jie He, Man-Shu Wang, Bin Yang, and Yu-Bo Li.
\newblock The evolution of single-cell rna sequencing technology and application: progress and perspectives.
\newblock {\em International Journal of Molecular Sciences}, 24(3):2943, 2023.

\bibitem{marguerat2010rna}
Samuel Marguerat and J{\"u}rg B{\"a}hler.
\newblock Rna-seq: from technology to biology.
\newblock {\em Cellular and molecular life sciences}, 67:569--579, 2010.

\bibitem{marouf2020realistic}
Mohamed Marouf, Pierre Machart, Vikas Bansal, Christoph Kilian, Daniel~S Magruder, Christian~F Krebs, and Stefan Bonn.
\newblock Realistic in silico generation and augmentation of single-cell rna-seq data using generative adversarial networks.
\newblock {\em Nature communications}, 11(1):166, 2020.

\bibitem{goodfellow2020generative}
Ian Goodfellow, Jean Pouget-Abadie, Mehdi Mirza, Bing Xu, David Warde-Farley, Sherjil Ozair, Aaron Courville, and Yoshua Bengio.
\newblock Generative adversarial networks.
\newblock {\em Communications of the ACM}, 63(11):139--144, 2020.

\bibitem{xu2020scigans}
Yungang Xu, Zhigang Zhang, Lei You, Jiajia Liu, Zhiwei Fan, and Xiaobo Zhou.
\newblock scigans: single-cell rna-seq imputation using generative adversarial networks.
\newblock {\em Nucleic acids research}, 48(15):e85--e85, 2020.

\bibitem{huang2023scggan}
Zimo Huang, Jun Wang, Xudong Lu, Azlan Mohd~Zain, and Guoxian Yu.
\newblock scggan: single-cell rna-seq imputation by graph-based generative adversarial network.
\newblock {\em Briefings in bioinformatics}, 24(2):bbad040, 2023.

\bibitem{zinati2024groundgan}
Yazdan Zinati, Abdulrahman Takiddeen, and Amin Emad.
\newblock Groundgan: Grn-guided simulation of single-cell rna-seq data using causal generative adversarial networks.
\newblock {\em Nature Communications}, 15(1):4055, 2024.

\bibitem{kingma2022autoencodingvariationalbayes}
Diederik~P Kingma and Max Welling.
\newblock Auto-encoding variational bayes, 2022.

\bibitem{svensson2020interpretable}
Valentine Svensson, Adam Gayoso, Nir Yosef, and Lior Pachter.
\newblock Interpretable factor models of single-cell rna-seq via variational autoencoders.
\newblock {\em Bioinformatics}, 36(11):3418--3421, 2020.

\bibitem{gronbech2020scvae}
Christopher~Heje Gr{\o}nbech, Maximillian~Fornitz Vording, Pascal~N Timshel, Casper~Kaae S{\o}nderby, Tune~H Pers, and Ole Winther.
\newblock scvae: variational auto-encoders for single-cell gene expression data.
\newblock {\em Bioinformatics}, 36(16):4415--4422, 2020.

\bibitem{wang2018vasc}
Dongfang Wang and Jin Gu.
\newblock Vasc: dimension reduction and visualization of single-cell rna-seq data by deep variational autoencoder.
\newblock {\em Genomics, Proteomics and Bioinformatics}, 16(5):320--331, 2018.

\bibitem{seninge2021vega}
Lucas Seninge, Ioannis Anastopoulos, Hongxu Ding, and Joshua Stuart.
\newblock Vega is an interpretable generative model for inferring biological network activity in single-cell transcriptomics.
\newblock {\em Nature communications}, 12(1):5684, 2021.

\bibitem{ho2020denoising}
Jonathan Ho, Ajay Jain, and Pieter Abbeel.
\newblock {Denoising diffusion probabilistic models}.
\newblock {\em Adv. Neural Inf. Process. Syst.}, 33:6840--6851, 2020.

\bibitem{dhariwal2021diffusion}
Prafulla Dhariwal and Alexander Nichol.
\newblock Diffusion models beat gans on image synthesis.
\newblock {\em Advances in neural information processing systems}, 34:8780--8794, 2021.

\bibitem{croitoru2023diffusion}
Florinel-Alin Croitoru, Vlad Hondru, Radu~Tudor Ionescu, and Mubarak Shah.
\newblock Diffusion models in vision: A survey.
\newblock {\em IEEE Transactions on Pattern Analysis and Machine Intelligence}, 45(9):10850--10869, 2023.

\bibitem{chen2023diffusiondet}
Shoufa Chen, Peize Sun, Yibing Song, and Ping Luo.
\newblock Diffusiondet: Diffusion model for object detection.
\newblock In {\em Proceedings of the IEEE/CVF international conference on computer vision}, pages 19830--19843, 2023.

\bibitem{yang2023diffusion}
Ling Yang, Zhilong Zhang, Yang Song, Shenda Hong, Runsheng Xu, Yue Zhao, Wentao Zhang, Bin Cui, and Ming-Hsuan Yang.
\newblock Diffusion models: A comprehensive survey of methods and applications.
\newblock {\em ACM Computing Surveys}, 56(4):1--39, 2023.

\bibitem{guo2023diffusion}
Zhiye Guo, Jian Liu, Yanli Wang, Mengrui Chen, Duolin Wang, Dong Xu, and Jianlin Cheng.
\newblock Diffusion models in bioinformatics: A new wave of deep learning revolution in action.
\newblock {\em arXiv preprint arXiv:2302.10907}, 2023.

\bibitem{watson2023novo}
Joseph~L Watson, David Juergens, Nathaniel~R Bennett, Brian~L Trippe, Jason Yim, Helen~E Eisenach, Woody Ahern, Andrew~J Borst, Robert~J Ragotte, Lukas~F Milles, et~al.
\newblock De novo design of protein structure and function with rfdiffusion.
\newblock {\em Nature}, 620(7976):1089--1100, 2023.

\bibitem{dong2024scrdit}
Shengze Dong, Zhuorui Cui, Ding Liu, and Jinzhi Lei.
\newblock scrdit: Generating single-cell rna-seq data by diffusion transformers and accelerating sampling.
\newblock {\em arXiv preprint arXiv:2404.06153}, 2024.

\bibitem{ronneberger2015u}
Olaf Ronneberger, Philipp Fischer, and Thomas Brox.
\newblock U-net: Convolutional networks for biomedical image segmentation.
\newblock In {\em Medical image computing and computer-assisted intervention--MICCAI 2015: 18th international conference, Munich, Germany, October 5-9, 2015, proceedings, part III 18}, pages 234--241. Springer, 2015.

\bibitem{peebles2023scalable}
William Peebles and Saining Xie.
\newblock Scalable diffusion models with transformers.
\newblock In {\em Proceedings of the IEEE/CVF International Conference on Computer Vision}, pages 4195--4205, 2023.

\bibitem{gao2023masked}
Shanghua Gao, Pan Zhou, Ming-Ming Cheng, and Shuicheng Yan.
\newblock Masked diffusion transformer is a strong image synthesizer.
\newblock In {\em Proceedings of the IEEE/CVF International Conference on Computer Vision}, pages 23164--23173, 2023.

\bibitem{yu2024white}
Yaodong Yu, Sam Buchanan, Druv Pai, Tianzhe Chu, Ziyang Wu, Shengbang Tong, Benjamin Haeffele, and Yi~Ma.
\newblock White-box transformers via sparse rate reduction.
\newblock {\em Advances in Neural Information Processing Systems}, 36, 2024.

\bibitem{yu2023white}
Yaodong Yu, Sam Buchanan, Druv Pai, Tianzhe Chu, Ziyang Wu, Shengbang Tong, Hao Bai, Yuexiang Zhai, Benjamin~D Haeffele, and Yi~Ma.
\newblock White-box transformers via sparse rate reduction: Compression is all there is?
\newblock {\em arXiv preprint arXiv:2311.13110}, 2023.

\bibitem{yu2020learningdiversediscriminativerepresentations}
Yaodong Yu, Kwan Ho~Ryan Chan, Chong You, Chaobing Song, and Yi~Ma.
\newblock Learning diverse and discriminative representations via the principle of maximal coding rate reduction, 2020.

\bibitem{chan2020deep}
Kwan Ho~Ryan Chan, Yaodong Yu, Chong You, Haozhi Qi, John Wright, and Yi~Ma.
\newblock Deep networks from the principle of rate reduction.
\newblock {\em arXiv preprint arXiv:2010.14765}, 2020.

\bibitem{chan2022redunet}
Kwan Ho~Ryan Chan, Yaodong Yu, Chong You, Haozhi Qi, John Wright, and Yi~Ma.
\newblock Redunet: A white-box deep network from the principle of maximizing rate reduction.
\newblock {\em Journal of machine learning research}, 23(114):1--103, 2022.

\bibitem{dosovitskiy2020image}
Alexey Dosovitskiy.
\newblock An image is worth 16x16 words: Transformers for image recognition at scale.
\newblock {\em arXiv preprint arXiv:2010.11929}, 2020.

\bibitem{he2022masked}
Kaiming He, Xinlei Chen, Saining Xie, Yanghao Li, Piotr Doll{\'a}r, and Ross Girshick.
\newblock Masked autoencoders are scalable vision learners.
\newblock In {\em Proceedings of the IEEE/CVF conference on computer vision and pattern recognition}, pages 16000--16009, 2022.

\bibitem{zhang2022dino}
Hao Zhang, Feng Li, Shilong Liu, Lei Zhang, Hang Su, Jun Zhu, Lionel~M Ni, and Heung-Yeung Shum.
\newblock Dino: Detr with improved denoising anchor boxes for end-to-end object detection.
\newblock {\em arXiv preprint arXiv:2203.03605}, 2022.

\bibitem{devlin2018bert}
Jacob Devlin.
\newblock Bert: Pre-training of deep bidirectional transformers for language understanding.
\newblock {\em arXiv preprint arXiv:1810.04805}, 2018.

\bibitem{radford2019language}
Alec Radford, Jeffrey Wu, Rewon Child, David Luan, Dario Amodei, Ilya Sutskever, et~al.
\newblock Language models are unsupervised multitask learners.
\newblock {\em OpenAI blog}, 1(8):9, 2019.

\bibitem{van2008visualizing}
Laurens Van~der Maaten and Geoffrey Hinton.
\newblock Visualizing data using t-sne.
\newblock {\em Journal of machine learning research}, 9(11), 2008.

\bibitem{ma2007segmentation}
Yi~Ma, Harm Derksen, Wei Hong, and John Wright.
\newblock Segmentation of multivariate mixed data via lossy data coding and compression.
\newblock {\em IEEE transactions on pattern analysis and machine intelligence}, 29(9):1546--1562, 2007.

\bibitem{vaswani2023attentionneed}
Ashish Vaswani, Noam Shazeer, Niki Parmar, Jakob Uszkoreit, Llion Jones, Aidan~N. Gomez, Lukasz Kaiser, and Illia Polosukhin.
\newblock Attention is all you need, 2023.

\bibitem{beck2009fast}
Amir Beck and Marc Teboulle.
\newblock A fast iterative shrinkage-thresholding algorithm for linear inverse problems.
\newblock {\em SIAM journal on imaging sciences}, 2(1):183--202, 2009.

\bibitem{wright2022high}
John Wright and Yi~Ma.
\newblock {\em High-dimensional data analysis with low-dimensional models: Principles, computation, and applications}.
\newblock Cambridge University Press, 2022.

\bibitem{agarap2019deeplearningusingrectified}
Abien~Fred Agarap.
\newblock Deep learning using rectified linear units (relu), 2019.

\bibitem{puram2017single}
Sidharth~V Puram, Itay Tirosh, Anuraag~S Parikh, Anoop~P Patel, Keren Yizhak, Shawn Gillespie, Christopher Rodman, Christina~L Luo, Edmund~A Mroz, Kevin~S Emerick, et~al.
\newblock {Single-cell transcriptomic analysis of primary and metastatic tumor ecosystems in head and neck cancer}.
\newblock {\em Cell}, 171(7):1611--1624, 2017.

\bibitem{Shim2022}
Joonho Shim, Se~Jin Oh, Eunhye Yeo, Ji~Hye Park, Jai~Hee Bae, Seok-Hyung Kim, Dongyoun Lee, and Jong~Hee Lee.
\newblock {Integrated analysis of single-cell and spatial transcriptomics in keloids: highlights on fibrovascular interactions in keloid pathogenesis}.
\newblock {\em J. Invest. Dermatol.}, 142(8):2128--2139, 2022.

\bibitem{zhu2023single}
Qionghua Zhu, Xin Zhao, Yuanhang Zhang, Yanping Li, Shang Liu, Jingxuan Han, Zhiyuan Sun, Chunqing Wang, Daqi Deng, Shanshan Wang, et~al.
\newblock Single cell multi-omics reveal intra-cell-line heterogeneity across human cancer cell lines.
\newblock {\em Nature Communications}, 14(1):8170, 2023.

\bibitem{kleiveland2015peripheral}
Charlotte~R Kleiveland.
\newblock Peripheral blood mononuclear cells.
\newblock {\em The Impact of Food Bioactives on Health: in vitro and ex vivo models}, pages 161--167, 2015.

\bibitem{kullback1951information}
Solomon Kullback and Richard~A Leibler.
\newblock On information and sufficiency.
\newblock {\em The annals of mathematical statistics}, 22(1):79--86, 1951.

\bibitem{gretton2006kernel}
Arthur Gretton, Karsten Borgwardt, Malte Rasch, Bernhard Sch{\"o}lkopf, and Alex Smola.
\newblock A kernel method for the two-sample-problem.
\newblock {\em Advances in neural information processing systems}, 19, 2006.

\bibitem{song2020denoising}
Jiaming Song, Chenlin Meng, and Stefano Ermon.
\newblock Denoising diffusion implicit models.
\newblock {\em arXiv preprint arXiv:2010.02502}, 2020.

\end{thebibliography}

\end{document}